\documentclass[conference]{IEEEtran}
\usepackage{times}
\usepackage[numbers]{natbib}
\usepackage{multicol}
\usepackage[bookmarks=true]{hyperref}
\usepackage{graphicx}
\usepackage[export]{adjustbox}
\usepackage{subcaption}
\usepackage{wrapfig}
\usepackage[numbers]{natbib}
\usepackage{multicol}
\usepackage[bookmarks=true]{hyperref}
\usepackage{booktabs}
\usepackage{graphicx}
\usepackage[linesnumbered,ruled]{algorithm2e}
\usepackage{float}
\usepackage{amsmath, amssymb}
\usepackage{amsfonts}
\usepackage{caption}
\newcommand{\shrinka}{\def\baselinestretch{0.993}\large\normalsize}

\begin{document}

\title{DeformerNet: A Deep Learning Approach to 3D Deformable Object Manipulation}

% You will get a Paper-ID when submitting a pdf file to the conference system
\author{Bao Thach*, Alan Kuntz*, Tucker Hermans*\(^\dagger\)\\
*University of Utah Robotics Center;
\(^\dagger\)NVIDIA}

%\author{\authorblockN{Michael Shell}
%\authorblockA{School of Electrical and\\Computer Engineering\\
%Georgia Institute of Technology\\
%Atlanta, Georgia 30332--0250\\
%Email: mshell@ece.gatech.edu}
%\and
%\authorblockA{Twentieth Century Fox\\
%\and
%\authorblockN{James Kirk\\ and Montgomery Scott}
%\authorblockA{Starfleet Academy\\
%San Francisco, California 96678-2391\\

% avoiding spaces at the end of the author lines is not a problem with
% conference papers because we don't use \thanks or \IEEEmembership

%Homer Simpson\authorrefmark{2},
%Montgomery Scott\authorrefmark{3} and
%Eldon Tyrell\authorrefmark{4}}
%Telephone: (800) 555--1212, Fax: (888) 555--1212}
%\authorblockA{\authorrefmark{4}Tyrell Inc., 123 Replicant Street, Los Angeles, California 90210--4321}}

\maketitle

% \begin{abstract}
% In this paper, we propose a novel approach to 3D deformable object manipulation leveraging a deep neural network (DNN) called DeformerNet. Controlling the shape of a 3D object requires a state representation that can capture the full 3D geometry of the object. Current methods work around this problem by defining a set of points on the object or only deform the object in 2D image space. Instead, we explicitly use 3D point clouds as the state representation and apply Covolutional Neural Network (CNN) to learn the 3D features important to the shape servo problem. These features are then mapped to robotic end-effector’s position using a multi-layer neural network. Once trained in an end-to-end fashion, DeformerNet directly maps the point clouds of the deformable object to a desired gripper position of the robot. In addition, we investigate the problem of detecting the manipulation point location.
% % by using unsupervised keypoint detection on point cloud. 
% \end{abstract}
\shrinka
\IEEEpeerreviewmaketitle
\shrinka
% \begin{figure}[th!]
%     \centering
%     \includegraphics[width=\textwidth]{figures/DeformNet_new.png} \\
%     % \includegraphics[width=\linewidth]{figures/feature_extract.png}
%     \caption{(Top) Architecture of DeformerNet; (Bottom) architecture of the feature extraction module.}
%     \label{fig:DeformerNet}\vspace{-20pt}
% \end{figure}
\section{Introduction}
% because 1) deformable objects have an infinite degree of freedom and 2) finding a state representation of these objects is difficult. 

Following Navarro-Alarcon \emph{et al.}~\cite{Navarro-Alarcon2017}, we adopt the terminology \textit{shape servo} to describe the manipulation task in which a robot manipulates a 3D deformable object into a desired shape. While performing this servo control, the robot estimates the state of the object with visual sensors and uses this as a feedback signal. However, the biggest question is: How do we obtain a good state representation of the object to inform the robot about the object’s shape? \emph{3D} in this context refers to \emph{triparametric} or \emph{volumetric} objects \cite{Sanchez2018robotic} which have no dimension significantly smaller than the other two, unlike \emph{uniparametric} (rope) and \emph{biparametric} objects (cloth).

A series of papers \cite{Navarro-Alarcon2013, Navarro-Alarcon2014, Navarro-Alarcon2016}  define a set of \textit{feature points} on the object as the state representation. These methods only work for known objects with distinct texture and cannot generalize to a diverse set of objects. This formulation also simply controls the displacements of individual points which does not fully reflect the 3D shape of the object. For precise control, one must use a large number of feature points, making control highly susceptible to noise and occlusion. \citet{Navarro-Alarcon2017} and \citet{Qi2019Contour} leverage 2D image contours as the feedback representations. However, using only 2D data severely limits the space of deformation control in 3D. 

\citet{Hu20193-D} address these shortcomings by using point clouds. They use extended FPFH \cite{Rusu2010VFH} to extract a feature vector from the point cloud and learn the deformation function via a Deep
Neural Network (DNN). We show that this hard-coded feature descriptor fails to generalize well.

Previous learning-based methods such as \cite{Ma2021, Wu2020Learning, Yan2020Learning} focus on rope and cloth. These works, often operated in image space, are not quite relevant to 3D object manipulation problems. There are also physical differences between rope-cloth and 3D objects (e.g. a 3D elastic object like soft tissue will return to its initial shape after the robot releases it), which makes methods in one domain not applicable to the other. 

Therefore, we propose a novel deep learning approach to solving the 3D shape servo problem. Instead of using the hard-coded feature vector as in \cite{Hu20193-D}, we develop a DNN that takes point clouds of the deformable objects as the inputs and outputs feature vectors. In addition, we develop a DNN that maps these feature vectors to the desired end-effector’s Cartesian position. We train both the feature extraction and deformation control neural networks together in an end-to-end fashion. Once trained, given point clouds of the object's current and goal shapes, the robot computes the desired position of its gripper at every time step. Finally, we study the problem of predicting the manipulation point. We are the first to propose a solution to this problem for 3D shape servoing.

% Our main contributions are:

% \begin{itemize}
% \item To the best of our knowledge, we are the first paper to successfully apply CNN on the deformable object shape control problem. Our novel approach allows the robot to extract features from the entire object shape in 3D, instead of just using individual points or 2D images as in \cite{Navarro-Alarcon2013, Navarro-Alarcon2014, Navarro-Alarcon2016} and \cite{Navarro-Alarcon2017}. 

% \item We propose a novel DNN, trained in an end-to-end fashion, which directly maps the point clouds of the deformable object to the desired gripper position of the robot. Our DNN consists of two stages: the first stage utilizes CNN to map the point clouds to a compact 256-dimension feature vector, and the second employs a fully-connected neural network to map the feature vector to end-effector positions.

% \item We propose a learning-based feature extraction for shape servoing task which is superior to the current state-of-the-art FPFH method utilized in \cite{Hu20193-D} 
% \end{itemize}
%%% Local Variables:
%%% mode: latex
%%% TeX-master: "main"
%%% End:

%\section{Method}\label{section:method}
\section{Shape Servoing}
\begin{figure}[t]
    \centering
    \includegraphics[width=1\linewidth]{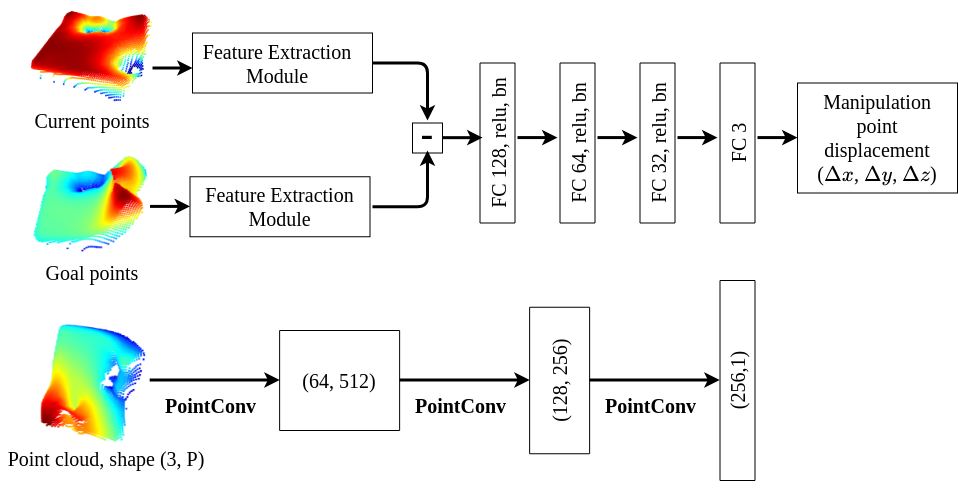} \\
    \caption{(Top) Architecture of DeformerNet; (Bottom) architecture of the feature extraction module.}
    \label{fig:DeformerNet}\vspace{-20pt}
\end{figure}
We define a set of manipulation points $p_m$ located on the surface of the deformable object where the robot gripper makes contact with the object and can directly control their positions. We formulate the shape servoing problem as finding a deformation function $F$ which maps the state representation $\psi$ of the deformable object to the manipulation points’ positions $p = [x, y, z] \in \mathbb{R}^{3}$. The intuition here is that we can drive the object to a desired shape by controlling the locations of these manipulation points: $p = F(\psi)$.
% \begin{equation}
% \label{deform_func_1}
%     p = F(\psi)
% \end{equation}

We choose $\psi$ to be a feature vector which encodes the shape of the deformable object.
We leverage a DNN derived from PointConv \cite{PointConv2019} which takes the point cloud of the current object shape and that of the goal object shape as the inputs and learns a 256-dimension feature vector. Figure~\ref{fig:DeformerNet} visualizes the network architecture of our feature extraction network.

For closed-loop operations, we modify the original equation to reflect the position and feature displacements:
%\begin{equation}
%\label{deform_func_2}
\(\Delta p = F(\Delta \psi)\)
%\end{equation}
% delta_p = F(delta_z),
where $\Delta \psi = \psi^* - \psi$ is the difference between the feature vector of the goal shape and that of the current object shape, and $\Delta p = p^* - p$ = $[\Delta x, \Delta y, \Delta z]^T$ is the relative displacement between the desired and current manipulation points. $\Delta p$ is also the Cartesian displacement of the robot end-effector since the robot directly controls the positions of the manipulation points.

We define this deformation function $F$ as a fully-connected neural network. The goal of the shape servoing problem then becomes learning a model which maps the feature vector displacement $\Delta \psi$ to the desired displacement of the manipulation points $\Delta p$. Given the learned $F$, we can now compute the desired end-effector position at every time step. The entire network architecture is shown in Fig.~\ref{fig:DeformerNet}. 
% The architecture consists of two stages: feature vector extraction and deformation control inference.

%%% Local Variables:
%%% mode: latex
%%% TeX-master: "main"
%%% End:

% \subsection{Runtime control strategy}
At runtime, the robot is given the point cloud of the current object shape and the point cloud of the goal object shape. First, we pass the point clouds through DeformerNet which outputs the desired gripper position. We then use the RRT-connect motion planner to plan a joint space path following the desired end-effector displacement. After the robot has reached an intermediate desired position, it gets a new point cloud of the current object shape and repeats the process.

% Instead of using the entire a plan, we only command the robot to follow first two intermediate configurations. 
% Our control strategy is is visualized in figure \ref{fig:control_strategy}

\begin{figure}[hb]
    \centering
    \includegraphics[width=\linewidth]{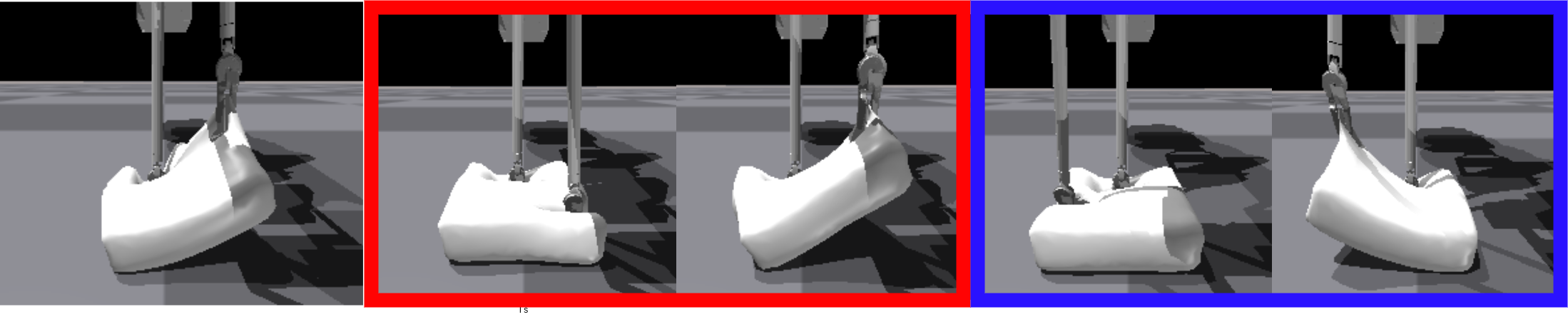} %height=80pt
    \caption{Importance of manipulation point (MP) selection. Leftmost: goal shape; Red: successful MP; Purple: failed MP.}
    \label{fig:mani_point}\vspace{-12pt}
\end{figure}

% \subsection{Manipulation point detection}
% \label{sec:mani_point_detection}
Prior works \cite{Hu20193-D, Navarro-Alarcon2013, Navarro-Alarcon2014, Navarro-Alarcon2016} assume that the manipulation points are given to the robot. However, a robot should select the best possible points to achieve the task. To understand the importance of selecting a good manipulation point, consider a simple scenario in Fig.~\ref{fig:mani_point}.
The leftmost image describes the goal shape; if the robot grasps the object as in the second image, it can successfully deform the object to a shape very close to the goal shape (third image).
However, with the manipulation point shown in the fourth image, the shape servo task now becomes impossible (rightmost image). 

We formulate the manipulation point as a function of the visual representations of the object: $x_m = f(\phi_0, \phi_g)$,
% \begin{equation}
% \label{mani_1}
%     x_m = f(\phi_0, \phi_g)
% \end{equation}
where $\phi_0$ is the current point cloud and $\phi_g$ is that of the goal shape. We propose two methods for deriving this function $f$.
\subsubsection{Using Keypoint Detection Heuristic}
We use an unsupervised keypoint detection algorithm derived from the method in \cite{Tejas2019unsupervised} to identify a set of K keypoints on the point clouds $Y_0 = (u_{10}, ..., u_{k0})$ and $Y_g = (u_1g, ..., u_kg)$ $\in \mathbb{R}^{K\times3}$. We define $\delta_Y = (\|u_{1g}-u_{10}\|, ..., \|u_{kg}-u_{k0}\|)$ which measures the displacement of each keypoint from the initial to the goal point cloud. We estimate the manipulation point location as the weighted average of the top $M$ keypoints that displace the most, with weights equal to the displacements of the keypoints.
% : $x_m = \frac{\sum_{i\in S}\delta_i p_i}{\sum_{i\in S}\delta_i}$,
% \begin{equation}
% \label{mani_3}
%     x_m = \frac{\sum_{i\in S}\delta_i^t p_i}{\sum_{i\in S}\delta_i^t}
% \end{equation}
% where $S$ is the set of top $M$ keypoints that displace the most.

\subsubsection{Using PointConv}

We learn the function $f$ as a regression problem using the same architecture as DeformerNet (Fig. \ref{fig:DeformerNet}). We modify the model to output the Cartesian position of the manipulation point $(x,y,z)$ instead of displacement.

%  From left to right: goal shape, MP 1, result using MP 1, MP 2, result using MP 2

\section{Experiments and results}
% \begin{figure*}[t]
%     % \centering
%     \includegraphics[scale=0.3]{figures/setup.png}
%     \caption{Experiment setup}
%     \label{fig:init_setup}
% \end{figure*}
%\subsection{Experiment setup and data collection}
\label{section:setup_data_collection}
\begin{figure}[t]
    \centering
    \includegraphics[scale=0.1]{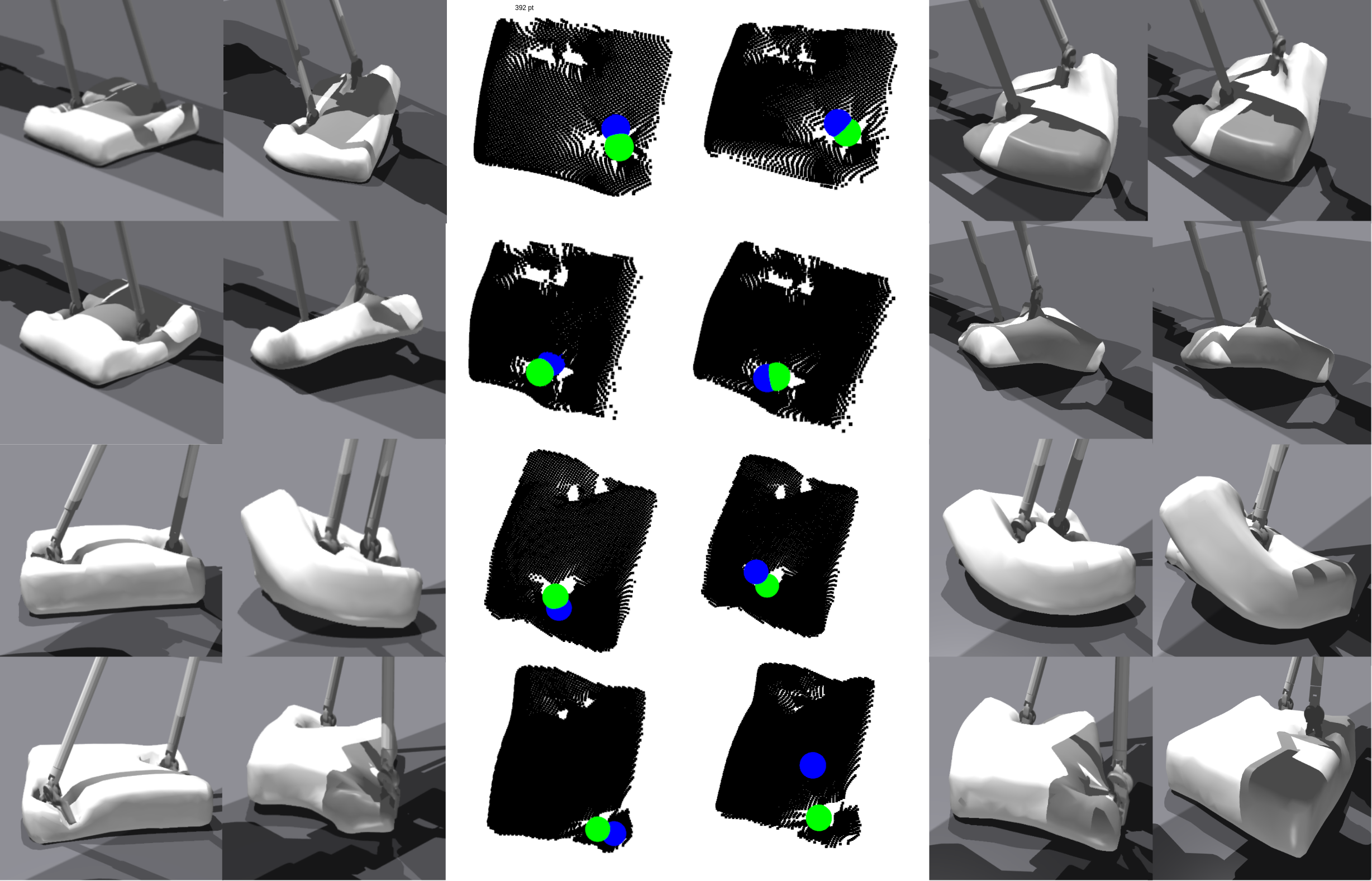}
    \caption{Left two columns: initial and goal shapes. Center columns: manipulation points predicted by keypoint method (left) and by PointConv (right); Blue spheres are the predicted MPs and green spheres are the ground truth. Right columns: shape servo results using MPs predicted in the center column.}
    \label{fig:shape_servo_result}
    \vspace{-12pt}
\end{figure}
In the Isaac Gym environment \cite{Liang2018GPU}, we use the bimanual daVinci surgical robot to manipulate a box object. One arm grasps one end of the object and holds it in place, while the other deforms the object into a desired shape. We created a training dataset by randomly sampling 150 pairs of object initial configurations and manipulation points. For each pair, the robot deforms the object to 200 random shapes.

Figure \ref{fig:shape_servo_result} shows the experimental results of our manipulation point selection algorithm using the two methods. We ran our Keypoint method and PointConv method with 100 test cases, and the average Euclidean distance between the predicted manipulation point and ground truth was 0.035[m] and 0.036[m], respectively. 

%\begin{figure}[t]
%    \centering
%    \includegraphics[scale=0.2]{figures/setup.png}
%    \caption{Experiment setup}
%    \label{fig:init_setup}
%\end{figure}

%  The heatmap describes the displacement of each keypoint and how likely it is a MP. The blue sphere is the predicted MP location and the green sphere is the ground truth. 
% Table \ref{Tab:mani_point} shows the average Euclidean distance between the predicted MP and ground truth over 100 runs.

% \begin{table}[ht]
% \caption{Distance between predicted MP and ground truth [m]}
% \centering
%   \begin{tabular}{|c|c|c|} \hline
%       & Keypoint method & PointConv method\\
%     \hline
%     % Case 1      & 0.17 & 0.18 \\ 
%     % Case 2      & 0.79 & 0.40 \\
%     % Case 3      & 0.63 & 0.53\\ 
%     % Case 4      & 0.42 & 0.53\\   
%     Average over 100 cases      & 0.035 & 0.036\\   \hline 
%   \end{tabular}
%   \label{Tab:mani_point}
  
% \end{table}

%\subsection{Performance of shape servo controller}
We evaluate the performance on 4 different pairs of initial-goal object shapes. Figure \ref{fig:shape_servo_result} visualizes the shape servo results. We use Chamfer distance as the evaluation metric to describe how close the final object point cloud is to the goal point cloud. Table \ref{Tab:chamfer_distance} shows the final Chamfer distance in each scenario, using the manipulation points predicted by the two methods. 

\begin{table}[ht]
\caption{Shape servo results in Chamfer distance [m]}
\centering
  \begin{tabular}{|c|c|c|} \hline
      & Keypoint method & PointConv method\\
    \hline
    Case 1      & 0.23 & 0.18 \\ 
    Case 2      & 0.32 & 0.39 \\
    Case 3      & 0.26 & 0.53\\ 
    Case 4      & 0.22 & 0.52\\   \hline  
  \end{tabular}
  \label{Tab:chamfer_distance}
\end{table}

We compare our DeformerNet with \cite{Hu20193-D}, the current state-of-the-art work for learning-based 3D shape servoing. We show that the hard-coded feature vector limits the deformation function model to only learn a small set of shapes. In contrast, using learned feature vectors, DeformerNet can fit a large number of shapes and hence outperforms the method from~\cite{Hu20193-D}.
% Hu \emph{et al.}'s

\citet{Hu20193-D}'s model underfits the data and results in very high train and test MSE losses (41.2 and 125.1 mm$^2$). Thus, the controller performs poorly and leads to very high final Chamfer distances ($>1$m) in all 4 test cases even with ground-truth manipulation point. In contrast, the losses of our DeformerNet are almost equal to zero (1.5 and 2.4 mm$^2$). Furthermore, when we train Hu \emph{et al.}'s model on a dataset with only a few shapes, the resulted MSE losses and final Chamfer distances become much smaller. This proves that while the previous method works well when trained on only a small set of shapes, it fails to generalize on our full dataset. 
\pagebreak

\bibliographystyle{plainnat}
\bibliography{references}

\clearpage
\section*{Appendix}
\subsection{Additional Experiment Visualizations}
\begin{figure*}[th!]
     \centering
     \begin{subfigure}[b]{\textwidth}
         \centering
         \includegraphics[width=1\textwidth]{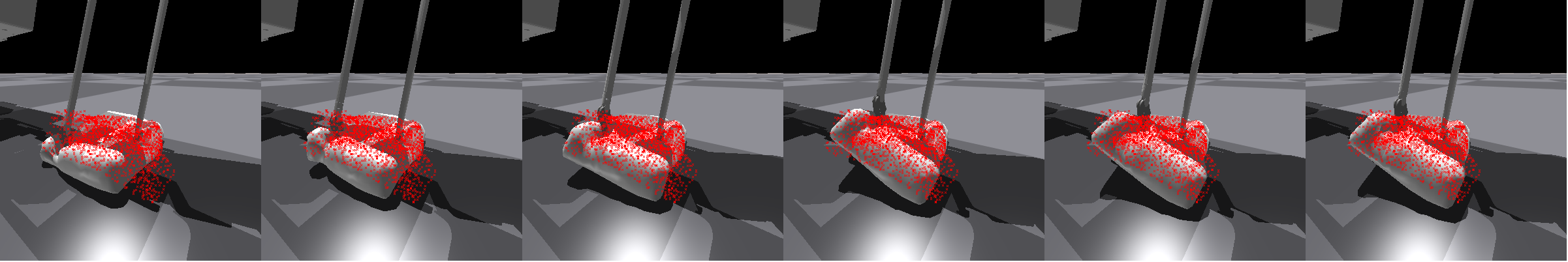}
         \caption{Case 1 shape servo sequence}
     \end{subfigure}
     \hfill
     \begin{subfigure}[b]{\textwidth}
         \centering
         \includegraphics[width=1\textwidth]{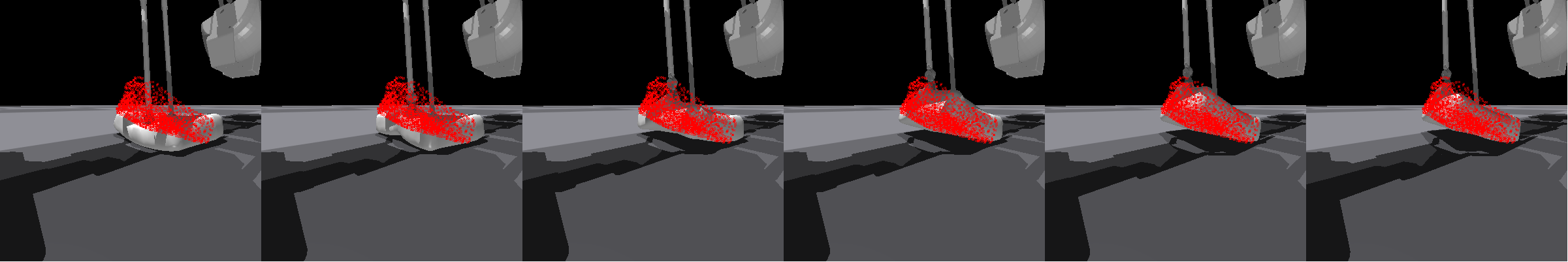}
         \caption{Case 2 shape servo sequence}
     \end{subfigure}
     \hfill     
     \begin{subfigure}[b]{\textwidth}
         \centering
         \includegraphics[width=1\textwidth]{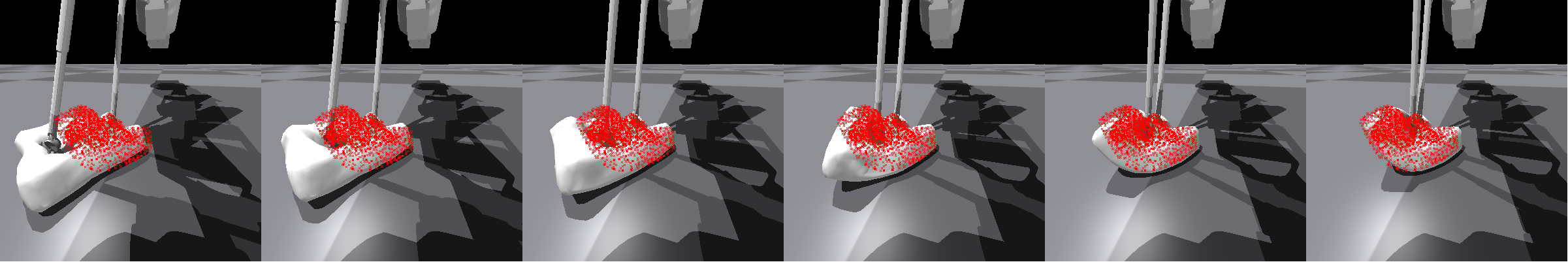}
         \caption{Case 3 shape servo sequence}
     \end{subfigure}
     \hfill
     \begin{subfigure}[b]{\textwidth}
         \centering
         \includegraphics[width=1\textwidth]{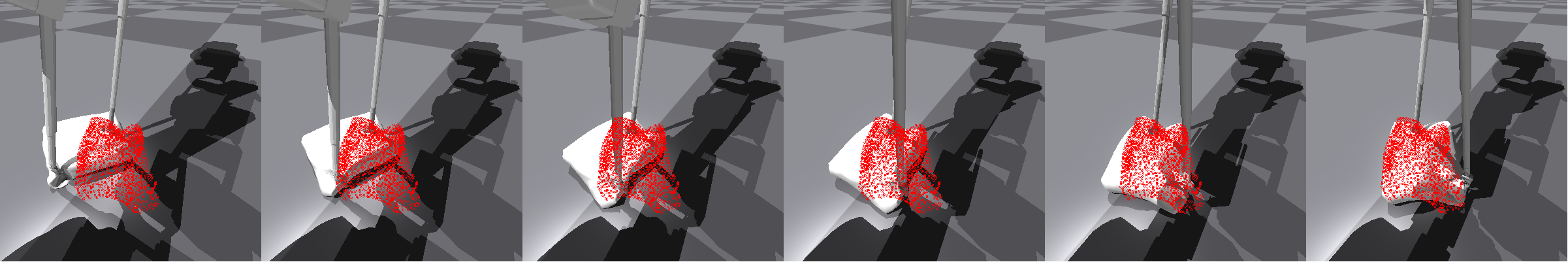}
         \caption{Case 4 shape servo sequence}
     \end{subfigure}
    \caption{Additional visualizations of the robot performing shape servoing to a variety of target shapes. The sparse red clouds visualize the target shapes of the object.}
    \label{fig:additional_vis}
\end{figure*}

\begin{figure*}[hb!]
    \centering
    \includegraphics[width=\linewidth]{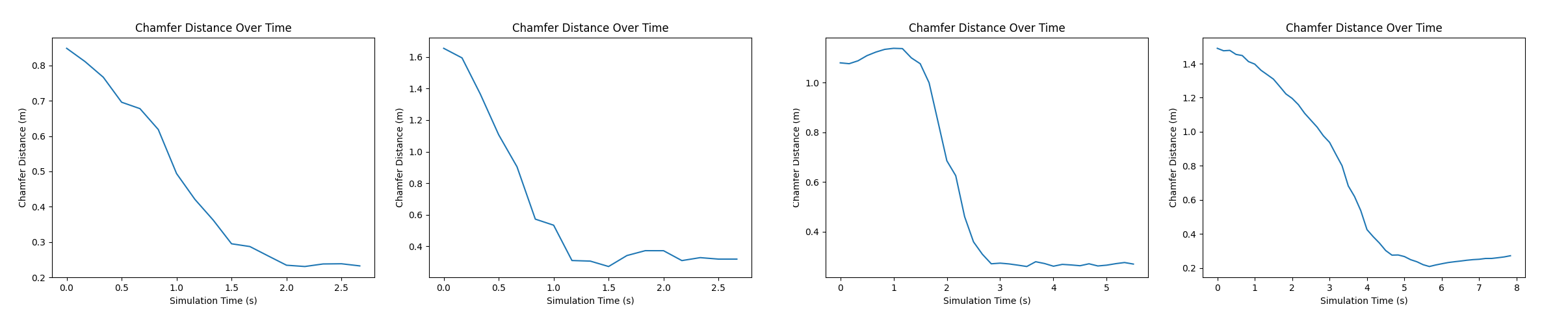}
    \caption{Chamfer distance between the current object point cloud and the goal object point cloud over time. From left to right: cases 1, 2, 3, \& 4 respectively.}
    \label{fig:chamfer_plots}
    \vspace{-12pt}
\end{figure*}

Due to space limits, we only present the final shape servo results in the main text. Here, for each shape servo test case mentioned in the experiment, we provide key frames of the whole manipulation sequence (Fig. \ref{fig:additional_vis}). In addition, Fig.~\ref{fig:chamfer_plots} shows the Chamfer distance between the current object point cloud and the goal object point cloud over time when running our DeformerNet controller. 

\begin{figure}[ht!]
    \centering
    \includegraphics[scale=0.3]{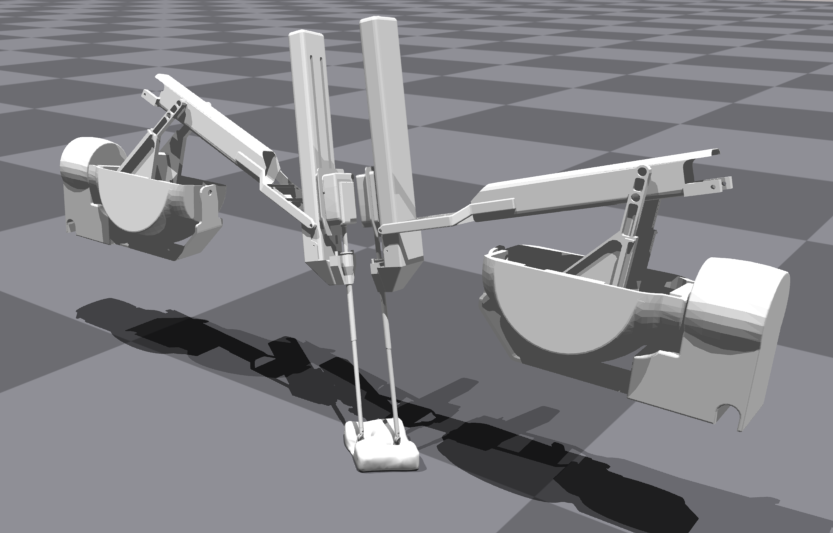}
    \caption{Experimental setup of the two armed daVinci surgical robot in the Issac gym simulator \cite{Liang2018GPU}.}
    \label{fig:init_setup}
    \vspace{-12pt}
\end{figure}

Figure \ref{fig:init_setup} shows the experimental setup. We use the bimanual daVinci surgical robot to manipulate a soft box object which mimics human tissue. The left arm grasps one end of the object and holds it in place, while the right arm deforms the object into a desired shape.

\subsection{DeformerNet  Details}
As shown in Fig.~\ref{fig:DeformerNet}, DeformerNet consists of two stages: feature vector extraction and defomation control inference. In the first stage, we perform convolution on 3D point clouds to extract feature vectors representing the states of the object shapes. This stage takes two point clouds as the inputs: one of the current object shape and one of the goal object shape. The output of this stage is two 256-dimension feature vectors. 

We then subtract the two feature vectors one from another and feed this to the second stage. The deformation control inference stage takes this 256-dimension \textit{differential feature vector} and passes it through a series of fully-connected layers (128, 64, and 32 neural units, respectively). The fully-connected output layer produces the 3D manipulation point displacement. Note that this is also equivalent to the robot’s end-effector Cartesian position displacement since the robot directly controls the position of the manipulation point. We use an ReLU activation function and batch normalization for all convolutional and fully-connected layers except for the output layer.

We use the standard mean squared error loss function for training our DNN. We adopt the Adam optimizer and a decaying learning rate which starts at $10^{-3}$ and decreases by 1/10 every 50 epochs. 

\subsection{Experiment Details}
Partial point clouds are generated and segmented out from the robot and background using the depth camera available inside the Issac gym environment. We sample 2048 points on each object point cloud using the Furthest Point Sampling method from \cite{Qi2017PointNet}. For the Keypoint Detection Heuristic method, we use 200 keypoints on each point cloud. The physical property of the object used in the experiment is: Young modulus = 1000 Pa, Poisson = 0.3.

\end{document}